%% file: main.tex
\documentclass{egpubl}
\usepackage{VMV2026}

\WsPaper           

\usepackage[T1]{fontenc}
\usepackage{dfadobe}

\usepackage{cite}  
\BibtexOrBiblatex
\electronicVersion
\PrintedOrElectronic

\ifpdf \usepackage[pdftex]{graphicx} \pdfcompresslevel=9
\else \usepackage[dvips]{graphicx} \fi

\usepackage{amsmath}
\usepackage{amssymb}
\DeclareMathOperator*{\argmin}{arg\,min}
\DeclareMathOperator*{\argmax}{arg\,max}
\usepackage{array}     
\usepackage{calc}      
\usepackage{multirow}  
\usepackage{tikz}      
\usetikzlibrary{calc}
\usetikzlibrary{arrows.meta}   
\usepackage{xcolor}    
\usepackage{booktabs}  
\usepackage{egweblnk}

\title[Partial Symmetry Detection via Contrastive Learning]%
      {Partial Symmetry Detection for 3D Geometry using\\
       Contrastive Learning with Geodesic Point Cloud Patches}

\author[G. Kobsik, I. Lim \& L. Kobbelt]
{\parbox{\textwidth}{\centering
  G.\,Kobsik, I.\,Lim and L.\,Kobbelt
  }\\
{\parbox{\textwidth}{\centering
  Visual Computing Institute, RWTH Aachen University, Germany
  }
}}

\begin{document}

\makeatletter
\def\ps@titlepage{\let\@mkboth\@gobbletwo
  \let\@oddhead\@empty \let\@evenhead\@empty
  \let\@oddfoot\@empty \let\@evenfoot\@empty}
\makeatother
\pagestyle{empty}

\setlength{\textfloatsep}{4pt}
\setlength{\intextsep}{4pt}
\setlength{\floatsep}{8pt}

\maketitle

\begin{abstract}
Detecting partial extrinsic symmetry in 3D geometry is a fundamental yet persistent challenge in computer vision and graphics, critical for tasks ranging from shape completion to procedural generation.
Classical transformation-space voting methods rely on pairwise matching, scaling as $O(n^2)$ and struggling to resolve coherent multi-instance groups.
Recent learning approaches advance global symmetry detection but restrict the solution space to reflection planes, failing to capture rotational or translational repetitions such as the legs of a chair or the steps of a staircase.
We propose SymCL, a self-supervised contrastive learning framework that detects partial symmetries across rotation, translation, and reflection (with scale-invariant features) and requires no ground truth annotations.
By mapping local geodesic patches to a latent space invariant to the Euclidean group, we reformulate symmetry detection as a density-based clustering problem, enabling the simultaneous discovery of multi-instance symmetric relationships in a single forward pass.
We evaluate quantitatively on SymPartNet, a new benchmark annotating all PartNet categories with partial symmetry relations, and demonstrate class-agnostic generalization qualitatively on everyday objects outside the training distribution.


\end{abstract}

\begin{figure}[t]
    \centering
    \includegraphics[width=\columnwidth]{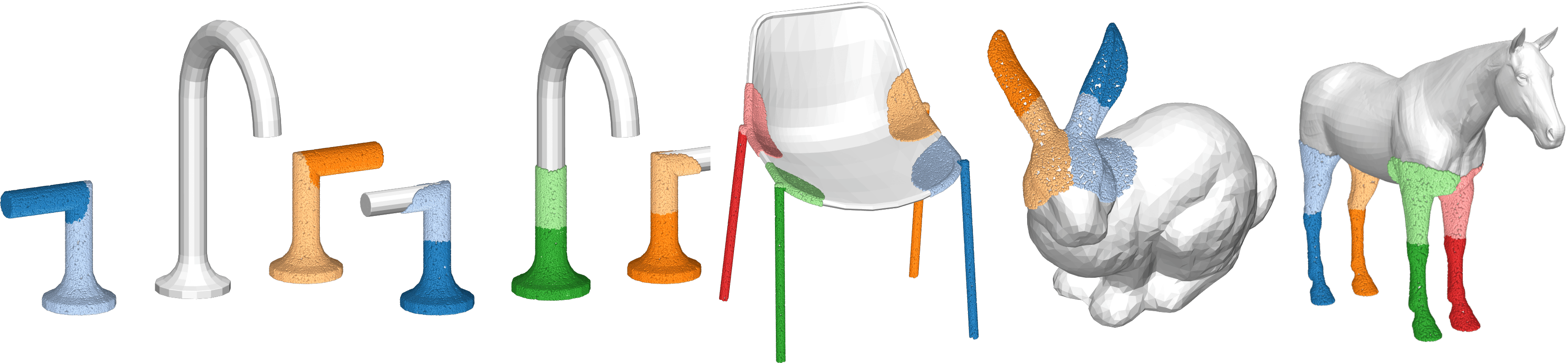}\\
    \includegraphics[width=\columnwidth]{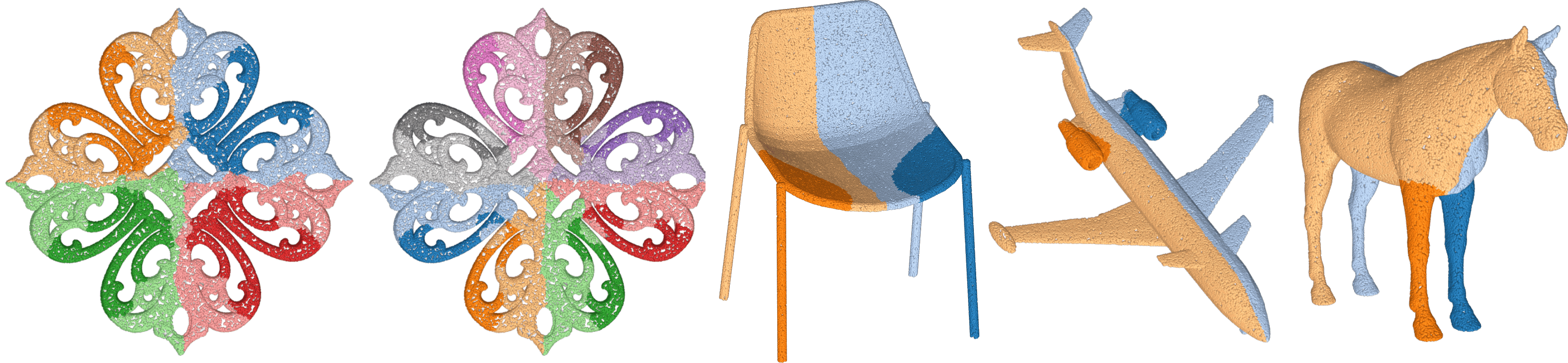}
    \caption{
        Detected symmetry groups after region growing.
        Colors represent distinct groups; dark shades are geodesic patch seeds, light shades are expanded regions.
        Multiple valid decompositions per shape are recovered simultaneously.
    }
    \label{fig:region_growing}
\end{figure}

\section{Introduction}
\label{sec:introduction}

Symmetry detection in 3D geometry is a challenging task that has been studied for many years.
The extraction of symmetry information is highly beneficial in the domain of computer vision and computer graphics for various downstream tasks, e.g.\ 3D geometry completion, segmentation, hierarchical decomposition, compression or shape matching, structure-aware meshing and procedural generation.
There are various solutions, including model-driven and learning-based methods, for global extrinsic symmetry detection.
However, research on partial extrinsic symmetry detection, also called self-similarity, has not received the same level of attention in recent years.
Note that the problem of detecting partial symmetries is ambiguous, since multiple valid solutions for almost any input can exist (see Fig.~\ref{fig:region_growing}).

Classical methods rely on hand-crafted features and model-driven algorithms, but are highly susceptible to parameter tuning and rarely generalize beyond specific shapes.
At their core, detecting symmetry between two regions $A$ and $B$ requires registration, distance computation, and thresholding, a pipeline that becomes intractable at scale: with $n$ candidate patches, all pairwise registrations cost $O(n^2)$ and each registration may take seconds.

Our aim is to find symmetric regions of maximal size within a shape.
To circumvent the time-consuming registration process we propose to learn rotation, reflection, translation and scale invariant features for local geodesic patches via contrastive learning, then cluster them in latent space.
The key assumption is that symmetric patches are geometrically congruent: under transformation invariance, geometric similarity becomes a reliable proxy for symmetry.
This formulation discovers all $n$ instances of a recurring structure as a single high-density cluster, bypassing the $O(n^2)$ cost of transformation-space voting.
We show that the learned features generalize to different geometry classes and unseen datasets without any adjustments.

A partial symmetry is naturally a multi-instance group: where a structure recurs $k$ times (e.g.\ the four legs of a chair or the eight arms of a filigree, see Fig.~\ref{fig:region_growing}), all $k$ parts are mutually symmetric, yet methods reasoning over isolated pairs recover individual correspondences rather than the group as a whole.
SymCL is \emph{amortized}: a single trained model recovers these multi-instance symmetry groups of arbitrary arity ($k$-fold rotational, translational, or reflective) in one forward pass, without per-shape optimization and class-agnostically.
Our main contributions are:
\begin{itemize}
    \item \textbf{Symmetry Detection as Clustering:} We recover all multi-instance symmetric groups (rotation, translation and reflection) simultaneously in a single forward pass by clustering $E(3)$-invariant geodesic-patch embeddings, in a self-supervised and class-agnostic manner, without committing to a symmetry type a priori.
    \item \textbf{SymPartNet Benchmark:} We provide, to our knowledge, the first benchmark for partial extrinsic symmetry detection, annotating PartNet with per-part symmetry relations and defining three retrieval metrics under two evaluation protocols.
\end{itemize}

\section{Related Work}
\label{sec:related_work}

Mitra et al.~\cite{mitra2013symmetry} survey classical model-based symmetry analysis in 3D geometry, noting that most organic and man-made objects exhibit some extractable symmetry.

\textbf{Global Symmetry.}
The most studied setting is global symmetry, where a single transformation maps the entire shape onto itself.
It has been studied long before in the 2D domain~\cite{atallah1985symmetry} and many classical algorithms have been proposed in 3D, predominantly for reflective planes based on Fourier descriptors~\cite{kazhdan2004reflective}, randomized voting techniques~\cite{podolak2006planar} or viewpoint entropy distribution~\cite{li2016efficient}, and also for rotational symmetries via generalized even moments~\cite{martinet2006accurate}.
Some of these approaches even aim at detecting global symmetry with missing geometry~\cite{sipiran2014approximate}.
In recent years novel learning-based techniques emerged, either supervised~\cite{ji2019fast} or unsupervised~\cite{gao2020prs, li2023e3sym}, and more recently single-image reflection detection~\cite{li2025symmetry}, though all are restricted to planar reflection.
Although these approaches surpass classical model-based methods and deal robustly even with incomplete data, they detect only global, not partial, symmetries.
Closest in spirit to ours, E3Sym~\cite{li2023e3sym} clusters $E(3)$-invariant point features to recover an arbitrary number of reflection planes; we likewise detect by clustering invariant embeddings, but target partial multi-instance symmetry of arbitrary type rather than global planar reflection alone.
A related but distinct family is \emph{intrinsic} symmetry, invariant to isometric (pose) deformations, addressed by classical global~\cite{ovsjanikov2008global, wang2017group} and partial~\cite{xu2009partial, mitra2010intrinsic, xu2012multi, nagar2018fast} methods as well as learning-based detection~\cite{qiao2022learning}.
As intrinsic symmetries are relevant only under isometric deformation, we focus on extrinsic partial symmetry, which covers rigid and articulated 3D models.

\textbf{Part Relationships.}
In shape encoding and generation~\cite{li2017grass, yu2019partnet, mo2019structurenet, gao2019sdm}, symmetric relationships are embedded into hierarchical shape representations to regularize the network, and can be recovered from the learned latent space by a separately trained point-cloud encoder~\cite{mo2019structurenet}.
However, such recovery remains limited to the training distribution and categories, whereas our method requires no labeled ground truth and generalizes to arbitrary geometry.

\textbf{Extrinsic Partial Symmetry.}
There exist classical model-based works that deal with the analysis of 3D shapes and the extraction of extrinsic partial symmetry, either based on geometric hashing~\cite{gal2006salient}, voting techniques in the transformation space~\cite{mitra2006partial, pauly2008discovering}, matching feature lines~\cite{bokeloh2009symmetry} or directly in the space of correspondences~\cite{lipman2010symmetry}.
Mitra et al.~\cite{mitra2013symmetry} abstract those techniques into three stages consisting of feature selection, aggregation and extraction.
Our work is inspired by this analysis, but aims at including the generalization capabilities of neural networks to improve upon them.
A key limitation of these classical methods is the sensitivity to manual parameter tuning and the lack of publicly available implementations.

Recently, Je et al.~\cite{je2024robust} propose an optimization-based approach that detects symmetries via per-shape score-field optimization using Riemannian Langevin dynamics.
Their method handles both partial and global reflective symmetry planes but requires per-shape optimization at inference time.
The authors outline a possible extension to rotational and translational symmetries but leave it to future work.
Architecturally, Je et al.\ follow a \emph{top-down} strategy: a global plane hypothesis is first fitted to the shape as a whole; supporting surface regions are derived post-hoc as a consequence of the detected plane.
In contrast, once trained our method recovers rotation, reflection and translation symmetries (with scale-invariant features) for arbitrary shape categories in a single forward pass, without any per-shape optimization.
SymCL is \emph{bottom-up}: local patch embeddings drive detection, with symmetry groups emerging from clustering without presupposing any symmetry type.

\textbf{Contrastive Learning.}
Self-supervised contrastive learning maps augmented views of the same instance close in feature space while separating different instances~\cite{chen2020simple}; we adopt its NT-Xent objective, a generalization of the N-pair loss~\cite{sohn2016improved}.
Closely related to our patch embeddings is the line of work on learned local 3D descriptors, which embed local neighborhoods such that geometrically similar regions match, including rotation-invariant, unsupervised formulations~\cite{deng2018ppffoldnet}; these target registration and matching rather than symmetry.
At the shape level, contrastive pre-training has further been applied to point clouds~\cite{xie2020pointcontrast, afham2022crosspoint}, predominantly to learn transferable features for downstream recognition.
In contrast, we employ contrastive learning not as a pre-training stage but as the core detection mechanism: by enforcing invariance to rigid transformations and scale, symmetric patches collapse to the same latent region, reducing symmetry detection to clustering.

\section{Learning Partial Extrinsic Symmetries}
\label{sec:methodology}

\begin{figure*}[ht]
    \centering
    \newcommand{\ovarrow}{\makebox[\linewidth][c]{\tikz{\draw[-{Triangle[length=2.2mm,width=2.4mm]},line width=3pt,darkgray] (0,0)--(0.02\textwidth,0);}}}
    \newcommand{\ovimg}[1]{\includegraphics[width=\linewidth]{img/overview/#1}}
    \newcommand{\ovimgs}[1]{\includegraphics[width=0.82\linewidth]{img/overview/#1}}
    \newcommand{\ovimgss}[1]{\includegraphics[width=0.738\linewidth]{img/overview/#1}}
    \setlength{\tabcolsep}{0pt}
    \newcolumntype{P}{>{\centering\arraybackslash}m{0.224\textwidth}}
    \newcolumntype{A}{>{\centering\arraybackslash}m{0.03\textwidth}}
    \begin{tabular}{@{}PAPAPAP@{}}
        \textbf{1) Feature Learning} & & \textbf{2) Feature Aggregation} & & \textbf{3) Symmetry Extraction} & & \textbf{4) Region Growing} \\
        \ovimg{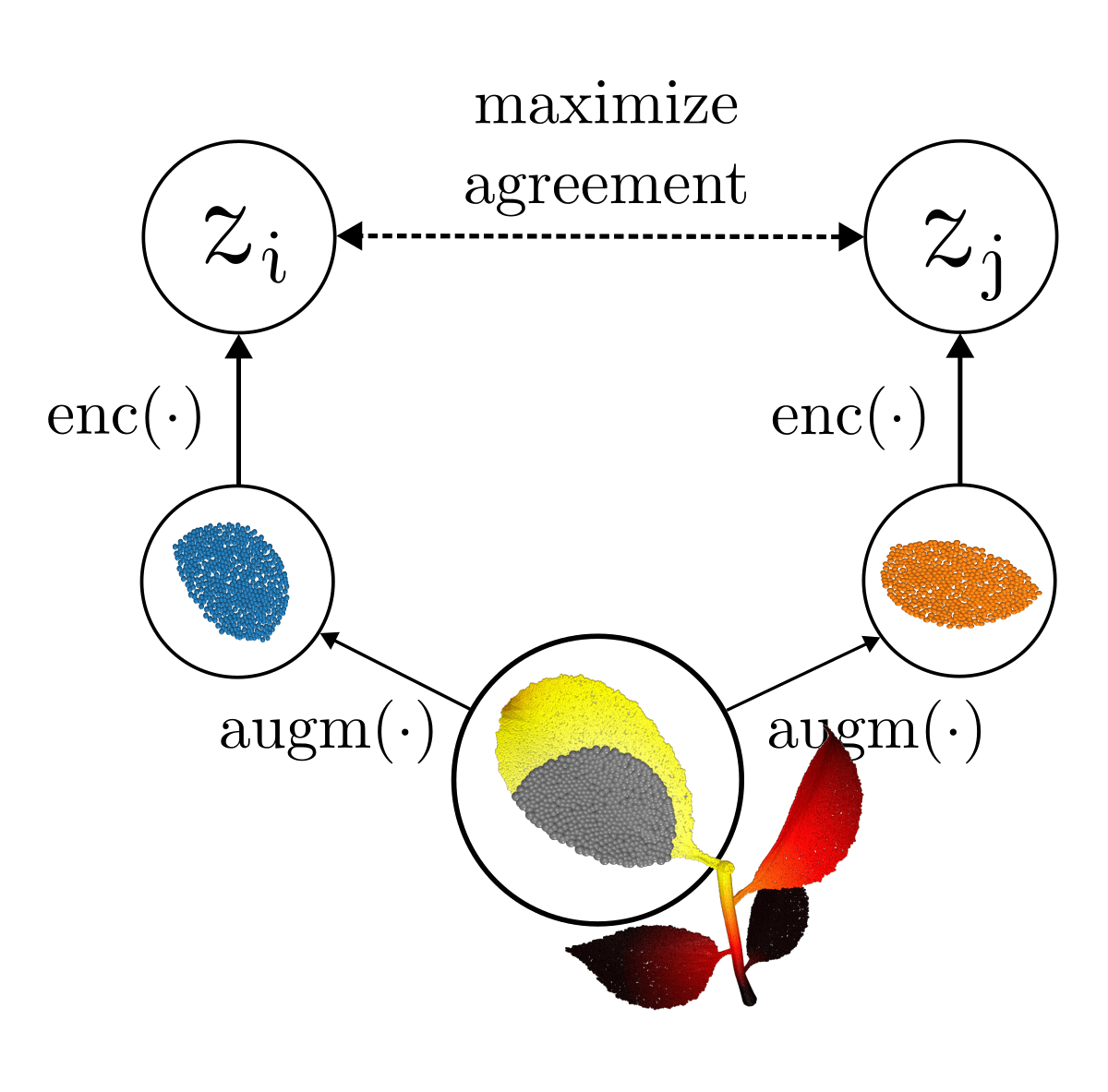} & \ovarrow &
        \ovimg{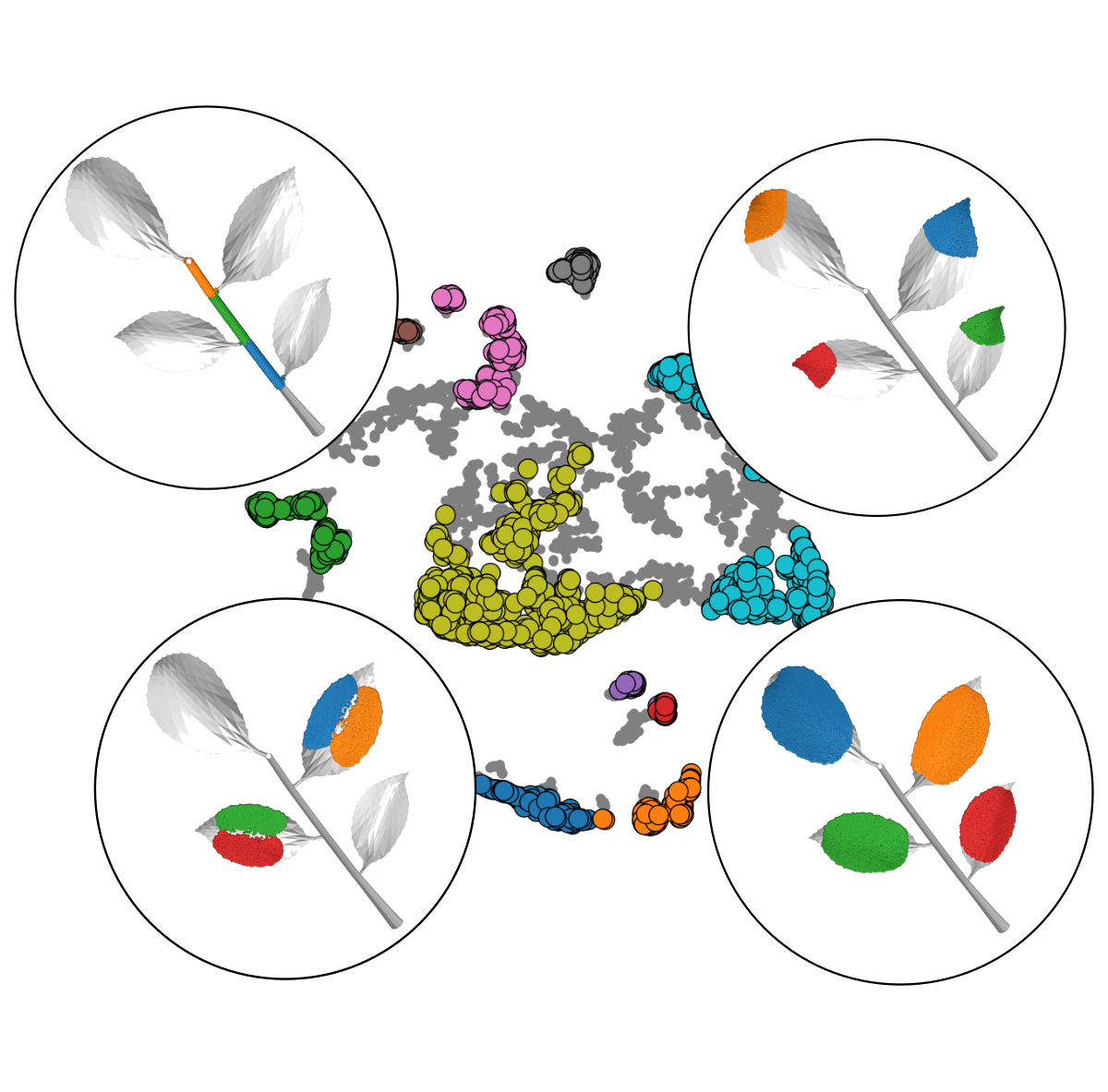} & \ovarrow &
        \ovimgs{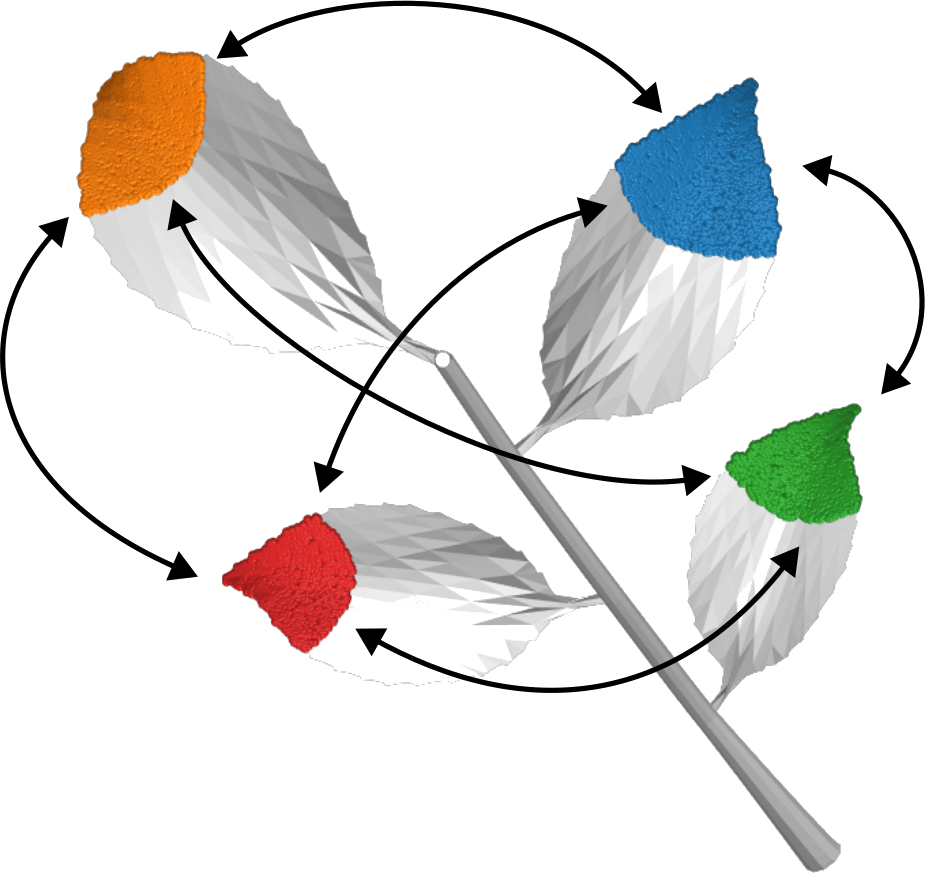} & \ovarrow &
        \ovimgss{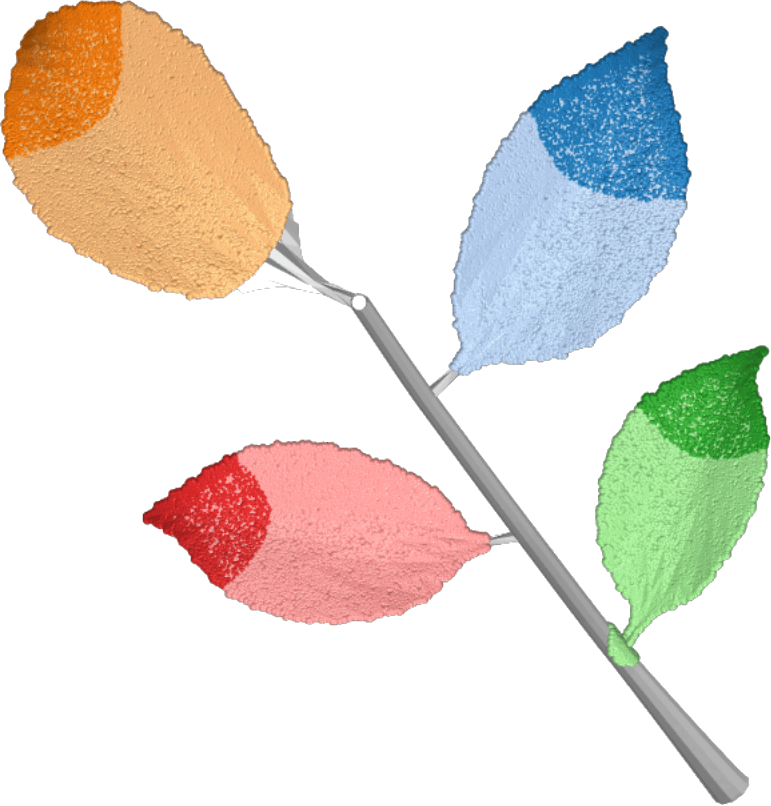} \\
    \end{tabular}
    \caption{
        Overview of our method to extract maximal regions of extrinsic partial symmetry.
        \textbf{1)~Feature Learning:} we learn features invariant to rotation, reflection, translation and scaling using geodesic patches.
        \textbf{2)~Feature Aggregation:} after training, we compare local patches in the latent space and aggregate symmetry information through clustering.
        \textbf{3)~Symmetry Extraction:} symmetries are extracted from the 3D geometry and their quality evaluated by the ICP distance.
        \textbf{4)~Region Growing:} we complete the symmetric regions by a region growing algorithm.
    }
    \label{fig:overview}
\end{figure*}

We present SymCL, a self-supervised contrastive framework for partial extrinsic symmetry detection.
Our detection pipeline consists of three core stages: 1)~\emph{invariant feature learning}, 2)~\emph{latent clustering} for multi-instance symmetry aggregation, and 3)~\emph{geometric verification and extraction} (see Fig.~\ref{fig:overview}).
An optional 4)~\emph{region growing} step then expands the detected seed regions into maximal symmetric regions covering the symmetric portions of the shape.
In the first stage we learn a feature encoder whose latent space is invariant to rigid transformations, such that symmetric patches (regardless of their position or orientation) map to the same region.
In the second stage, density-based clustering in latent space discovers all $n$ instances of a recurring structure simultaneously.
Lastly, we cast clusters back to 3D space, compute connected components, and verify geometric alignment via ICP distance.

\subsection{Geodesic Patches}
\label{sec:geodesic_patches}

To enable partial extrinsic symmetry computation, we must discretize the continuous surface of a 3D geometry into local patches.
Patch centers are selected via the farthest point sampling (FPS) of a point cloud of the input geometry.
Typically, a ball-query per sample point is performed to extract a local region of the geometry~\cite{qi2017pointnet++}.
We argue that the extraction method, which relies on Euclidean distances, is flawed as it fails to respect the intrinsic metric of the input surface.
To address this issue, we suggest calculating geodesic distances using the heat method~\cite{crane2017heat, geometrycentral} originating from the patch centers and identifying the closest neighbors within a fixed distance $\delta_d$ or a predefined number of points $\delta_n$.
This directly determines the size of the patch.
For each center, we extract geodesic patches at multiple scales to capture symmetries across a range of neighborhood sizes.

\subsection{ICP Distance}
\label{sec:icp_distance}

As previously mentioned, we may assess the symmetry of two shapes in relation to each other by initially registering them and subsequently calculating the distance between them.
We describe a distance measure for point clouds that is invariant to translation, uniform scaling, and the orthogonal group $O(3)$ (rotation and reflection), which we call the ICP distance.
Given two point clouds $A$ and $B$:
1)~Register $A$ onto $B$ using the iterative closest point (ICP) algorithm~\cite{besl1992method}.
2)~Compute the two-sided Chamfer distance $d_{CD}$~\cite{fan2017point} between the aligned point clouds $A_{ICP}$ and $B$.
\begin{equation}
    d_{CD}(A, B) =
    \frac{1}{|A|}\sum_{a \in A} \min_{b \in B} \lVert a - b \rVert^2 +
    \frac{1}{|B|}\sum_{b \in B} \min_{a \in A} \lVert a - b \rVert^2
\end{equation}

Before registration we normalize each point cloud to the unit sphere via $\mathcal{N}(\cdot)$, which mean-centers the cloud and divides it by its maximal radius from the centroid.
This removes translation and uniform scale, so that the ICP distance compares shape rather than absolute size or position.
As the ICP algorithm is highly sensitive to the initialization, we repeat the registration process several times, applying a random rotation and reflection $R_{rand}(\cdot)$ to $A$ each time, and define the smallest distance across all runs as the ICP distance $d_{ICP}$.
As a compromise between consistency and speed we employ $N = 30$ separate registrations, with a maximum of $100$ iteration steps each.
We additionally apply FPS $\mathrm{fps}(\cdot)$ to restrict each point cloud to $512$ points, which speeds up computation and makes distances comparable across inputs of different resolution.

\begin{equation}
    d_{ICP}(A, B) = \min_{i \in N} d_{CD}(A^{norm}_{ICP, i}, B^{norm}),
\end{equation}
where
\begin{equation}
    A^{norm} = \mathcal{N}(\mathrm{fps}(A)), \quad B^{norm} = \mathcal{N}(\mathrm{fps}(B))
\end{equation}
\begin{equation}
    R_i, T_i = \mathrm{ICP}(R_{rand}(A^{norm}), B^{norm})
\end{equation}
\begin{equation}
    A^{norm}_{ICP, i} = R_i T_i A^{norm}
\end{equation}

Given this explicit measure of extrinsic similarity, local patches, regions, or parts can now be compared to each other and a distance matrix can be calculated to describe a shape's symmetry landscape.

\subsection{Feature Learning}
\label{sec:feature_learning}

To extract local symmetric patterns we need the symmetry distance matrix between all patches.
The ICP distance is a reliable measure for a single pair, but comparing two patches takes seconds, making the $O(n^2)$ pairwise computation impractical for the dense patch sampling we require: a $16 \times 16$ matrix takes $35$ seconds, while a $1k \times 1k$ matrix takes almost $4$ hours.
We therefore replace the explicit registration with a learned embedding: each patch $p$ is lifted to a feature space in which similarity reduces to a single dot product.
To this end we train a neural network as a feature extractor $enc(p^{norm}) = z$, where $p^{norm}$ is the normalized input patch and $z \in \mathbb{R}^{32}$ the embedded feature vector.

\textbf{Feature Invariance.}
To account for various types of symmetry, we must calculate features that remain unaffected by rotation, reflection, translation and scale.
We achieve rotation-invariance by utilizing Vector Neurons~\cite{deng2021vector} at the core of our model, using VN-DGCNN as a lightweight feature encoder $enc(\cdot)$.
Translation-invariance is achieved by centering the mean input point cloud at zero; scale-invariance by scaling each input patch to the size of a unit sphere.
During training we randomly reflect the data along each axis to account for reflectional symmetry.

\textbf{Contrastive Learning (SymCL).}
We train the network SymCL to map similar patches $(z_i, z_j)$ close and dissimilar patches $(z_i, z_k)$ far away from each other in the latent space utilizing contrastive learning and the Normalized Temperature-scaled Cross Entropy Loss~\cite{sohn2016improved}:
\begin{equation}
    L_{Xent} = - \mathrm{log} \frac
    {\mathrm{exp}(\mathrm{sim}(z_i, z_j)/\tau)}
    {\sum_{k=1}^{2N}1_{[k \neq i]}\mathrm{exp}(\mathrm{sim}(z_i, z_k)/\tau)},
\end{equation}
with cosine-similarity $\mathrm{sim}(z_i, z_j) = z_i \cdot z_j / (\lVert z_i\rVert \lVert z_j \rVert)$,
where $N$ and $\tau$ represent the batch size and the temperature, respectively.
Similar patches are defined as a pair of augmented patches originating from the same patch center; dissimilar patches are all other patches within the batch.
This allows us to train the network in a self-supervised manner.

We apply augmentation techniques in the spirit of contrastive learning.
First, we apply anisotropic scaling to the data within a range of $[0.8, 1.2]$ for each axis, to account for small shape and curvature variations.
Second, we offset the patch center within a range of $[0.0, 0.1\cdot\delta_n]$ with respect to the geodesic distance of the original patch center.
This captures the local geometry more comprehensively and makes features robust to discretization artifacts.

\subsection{Symmetry Detection}
\label{sec:symmetry_detection}

To extract symmetry information from an input shape during inference, we perform the following steps: i)~feature selection, ii)~aggregation and iii)~symmetry extraction.

\textbf{Feature Selection.}
Patch centers are selected from the input geometry using FPS.
We extract a predefined number of geodesic patches, which are afterwards embedded into the latent space using the trained feature encoder $enc(\cdot)$.
We found that using at least $1000$ patches leads to good results.

\textbf{Aggregation.}
By clustering patches in a rotation, reflection, translation and scale invariant latent space we can aggregate patches exhibiting partial extrinsic symmetries.
Each cluster suggests a hypothesis $H$ consisting of multiple regions that are symmetric to each other.
Each region $r_{i} \in H$ is formed by merging several patches.
To merge the patches, we cast the clustered features $z_{j}$ back into the 3D space of the geometry and calculate connected components based on the geodesic distance between the extracted patch centers.
For clustering, we utilize HDBSCAN~\cite{mcinnes2017hdbscan}, a fast and reliable non-parametric clustering algorithm capable of detecting an a-priori unknown quantity of clusters as well as outliers.

\textbf{Symmetry Extraction.}
The latent space is only an approximation of the geometric characteristics, and therefore requires filtering.
We verify detected regions $r_{i}$ by calculating the ICP distance between all regions of a single hypothesis ${r_0, ..., r_n} \in H$.
The hypothesis is rejected if the greatest distance exceeds a predetermined similarity threshold:
\begin{equation}
    \delta_{sim} < \max_{r_i \in H, r_j \in H} d_{ICP}(r_i, r_j) .
\end{equation}
This threshold is fixed to $\delta_{sim} = 0.005$ in our method and is geometrically interpretable: since the ICP distance is a two-sided Chamfer distance between clouds normalized to the unit sphere, $0.005$ corresponds to a mean point-to-surface deviation on the order of $\sqrt{0.005} \approx 0.07$, i.e.\ a few percent of the shape extent.
We also discard a hypothesis $H$ if it produces a singular region $H = \{r_{0}\}$ or if more than $n > 30$ regions are detected, as these empirically do not represent meaningful symmetry groups.
This verification still incurs pairwise ICP comparisons, but only \emph{within} a single hypothesis: with at most $30$ regions per cluster it reduces to a small, bounded $O(k^2)$ computation per hypothesis, rather than the $O(n^2)$ registration over all $n$ sampled patches that transformation-space voting requires.

\subsection{Symmetry-aware Region Growing}
\label{sec:region_growing}

The identified symmetric regions $r_i$ consist of only local patches with limited size.
To complete the symmetric regions we apply a region growing algorithm that partitions the input geometry into maximally symmetric regions $r_i^{max}$.

\begin{figure}[t]
    \centering
    \includegraphics[width=\columnwidth]{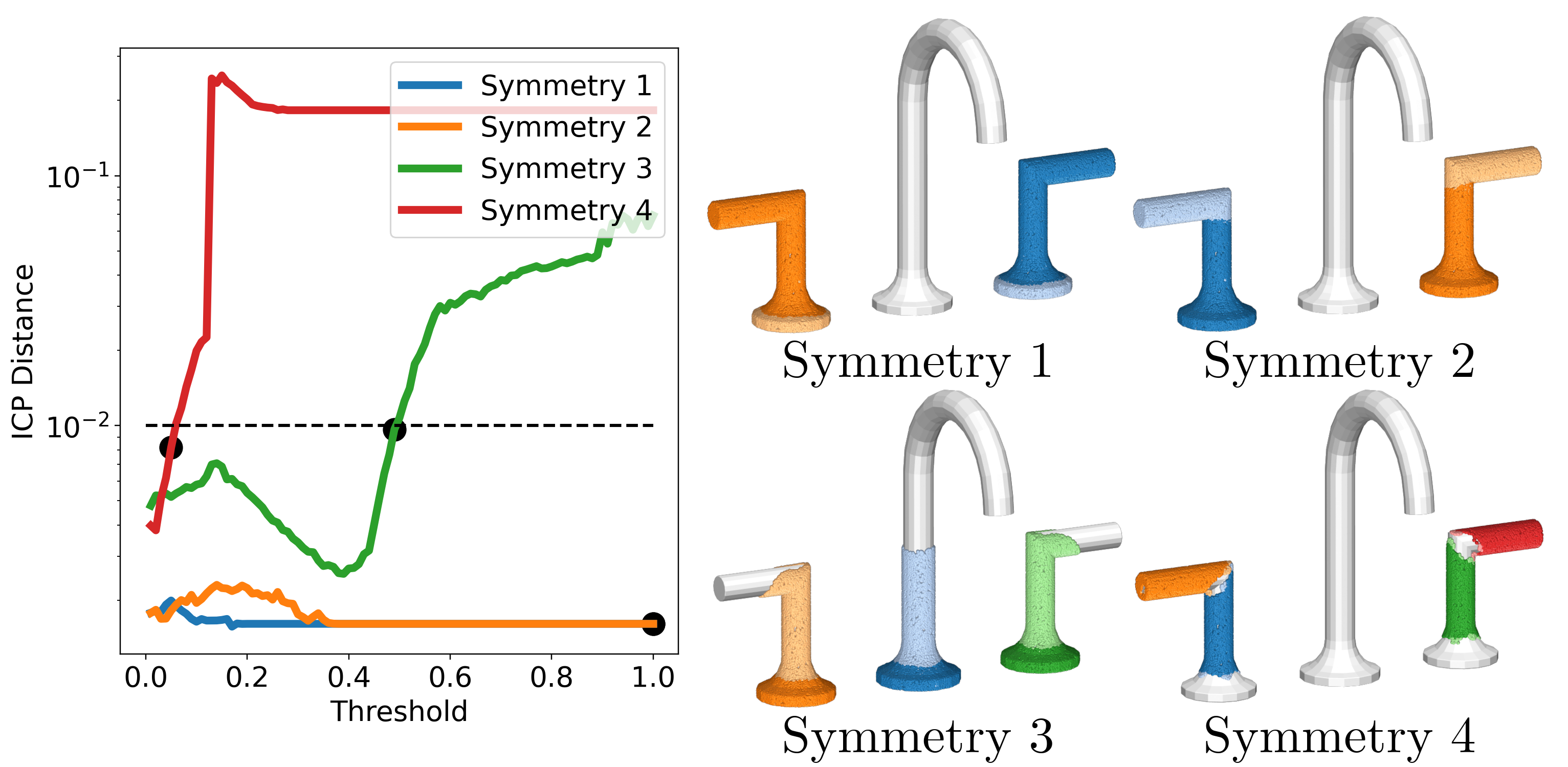}
    \caption{
        ICP distance during region growing for one detected symmetry pair.
        The region grows with increasing geodesic radius $\delta_g$ until the ICP distance rises sharply, marking the maximal geometrically valid symmetric partition.
    }
    \label{fig:region_growing_icp_dist}
\end{figure}

For each region $r_i \in H$ we calculate the geodesic distance from the region boundary to all other points.
Each point is assigned to the closest region $r_i$ according to its geodesic distance $d_{geo}$.
To identify the largest symmetric region $r_i^{max}$ we extract regions $r_i^{\delta_g}$, where each point satisfies $d_{geo} \leq \delta_g$, and compute the ICP distance between every pair.
We linearly interpolate the growing radius $\delta_g$ between $[0, d_{geo}^{max}]$ and select the largest symmetric regions $r_i^{max}$ (Fig.~\ref{fig:region_growing_icp_dist}) defined by:
\begin{equation}
    \delta_g^{max} = \argmin_{\delta_g \in [0, d_{geo}^{max}]} \max_{r_i \in H, r_j \in H} d_{ICP}(r_i^{\delta_g}, r_j^{\delta_g}).
\end{equation}

\section{SymPartNet Benchmark}
\label{sec:benchmark}

We could not identify any publicly available benchmark for the partial extrinsic symmetry \emph{detection} task; existing benchmarks target global reflective symmetry on synthetic point clouds~\cite{sipiran2023shrec}, and even recent work~\cite{je2024robust} notes the absence of a suitable dataset and resorts to ad-hoc annotation.
A symmetry-annotated PartNet exists~\cite{yu2019partnet}, but encodes a symmetry \emph{hierarchy} for structure learning, not a detection task with pairwise relations and metrics.
We annotate the PartNet dataset~\cite{mo2019partnet} with partial extrinsic symmetry relationships and establish two benchmarks.

\textbf{SymPartNet.}
We extend PartNet, comprising 26,671 semantically segmented 3D models across 24 object categories, with symmetry annotations at part-level for each model in the test set (SymPartNet.v1).
For each model we compute the ICP distance $d_{ICP}$ between all pairs of parts and threshold them at a manually chosen per-shape value $\delta_{sym}$ ($d_{ICP} \leq \delta_{sym}$).
This yields a binary part-symmetry matrix: an entry marks two parts as symmetric, i.e.\ mappable onto each other by a combination of rigid transformations and scaling.
The ground truth of each model is this matrix of part relationships, not the threshold itself.
Note that $\delta_{sym}$ is an annotation threshold, used only to define these ground-truth relations, and is distinct from the detection threshold $\delta_{sim}$ (Sec.~\ref{sec:symmetry_detection}), which our method applies at inference and keeps fixed across all shapes.

\textbf{Benchmarks.}
The \emph{SymPartNet (Parts)} protocol provides pre-segmented parts; the goal is to extract symmetric relations between them.
The \emph{SymPartNet (Full)} protocol provides only the raw 3D model; the goal is to compute partitions containing mutually symmetric regions.

\textbf{Evaluation Metrics.}
\label{sec:metrics}
Let $S_P = \{H_0, \dots, H_n\}$ be the predicted symmetry set and $S_{GT}$ the ground truth.

\textbf{ICP} measures geometric alignment within each predicted cluster:
\begin{equation}
    \text{ICP} =
    \frac{1}{|S_{P}|}
    \sum_{s_k \in S_{P}} \max_{r_i, r_j \in s_k} d_{ICP}(r_i, r_j).
\end{equation}

\textbf{IoU} measures overlap with ground truth clusters:
\begin{equation}
    \text{IoU} =
    \frac{1}{|S_{P}|}
    \sum_{s_k \in S_{P}} \max_{s_j \in S_{GT}} \frac{s_j \cap s_k}{s_j \cup s_k}.
\end{equation}

\textbf{COV} measures the fraction of ground truth symmetries covered:
\begin{equation}
    \text{COV} =
    \frac{1}{|S_{GT}|}
    |\{ \argmax_{s_j \in S_{GT}} \frac{s_j \cap s_k}{s_j \cup s_k} \}|,
    \quad \forall s_k \in S_P,
\end{equation}
where each ground truth cluster is counted at most once.

\section{Evaluation}
\label{sec:evaluation}

We train SymCL once on the PartNet~\cite{mo2019partnet} training split; all quantitative evaluation is performed on the held-out test split, and the qualitative results use shapes outside PartNet entirely.
Using the SymPartNet benchmark and metrics defined above, we evaluate SymCL in four experiments: (1) we verify that the learned latent space correctly organizes patches by geometric similarity, (2) we quantify detection quality on SymPartNet across all 24 PartNet categories, (3) we evaluate qualitatively on everyday objects to demonstrate class-agnostic generalization, and (4) we compare against competing methods on representative shapes spanning reflective, rotational, and translational structure.

\subsection{Latent Space Analysis}
\label{sec:latent_space}

The core claim of SymCL is that symmetric patches (regardless of their 3D position or orientation) map to the same high-density region in the learned latent space $\mathcal{Z}$.
We verify this directly before presenting downstream results.

\textbf{Computational Motivation.}
A brute-force ICP baseline for $N=5000$ patches requires a $5000{\times}5000$ symmetric distance matrix, ${\approx}48$ hours of pairwise registrations.
SymCL reduces this to $54$ seconds: one forward pass to embed all patches, followed by a single dot-product distance matrix computation, an $O(10^3)$ speedup that enables the dense sampling needed for fine-grained symmetry detection.

\begin{figure}[h]
    \centering
    {%
    \setlength{\tabcolsep}{1pt}
    \newcommand{\pw}{0.155\columnwidth}%
    \newcommand{\chairpw}{0.120\columnwidth}%
    \begin{tabular}{*{2}{>{\centering\arraybackslash}m{\pw}} @{\hspace{4pt}}!{\color{darkgray}\vrule}@{\hspace{4pt}} *{2}{>{\centering\arraybackslash}m{\pw}} @{\hspace{4pt}}!{\color{darkgray}\vrule}@{\hspace{4pt}} *{2}{>{\centering\arraybackslash}m{\pw}}}
        \multicolumn{2}{c}{Filigree} &
        \multicolumn{2}{c}{Chair} &
        \multicolumn{2}{c}{Faucet} \\[2pt]
        \includegraphics[width=\pw]{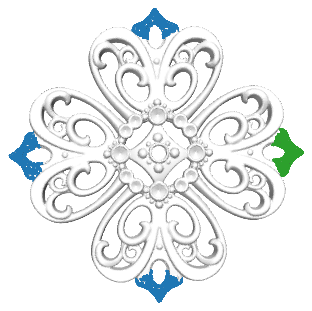} &
        \includegraphics[width=\pw]{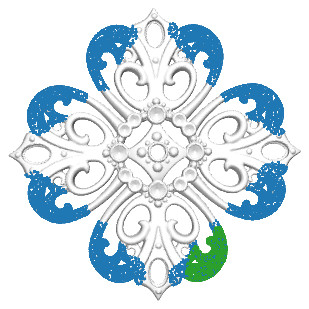} &
        \includegraphics[width=\chairpw]{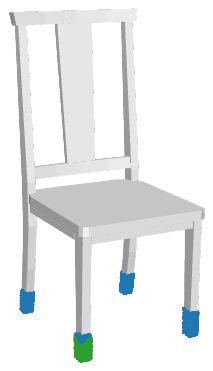} &
        \includegraphics[width=\chairpw]{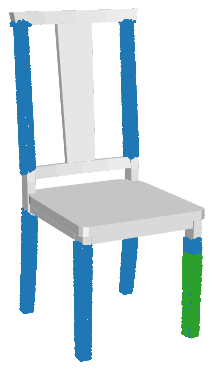} &
        \includegraphics[width=\pw]{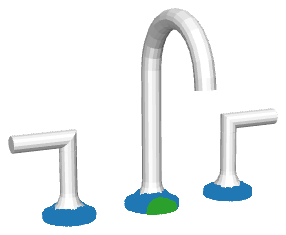} &
        \includegraphics[width=\pw]{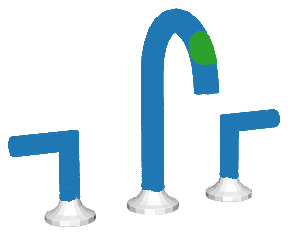} \\
    \end{tabular}
    }%
    \caption{
        Probe retrieval in the learned latent space.
        Selecting one patch (green) and coloring all others by cosine similarity reveals that similar patches (blue) cluster together irrespective of their spatial location.
    }
    \label{fig:probe}
\end{figure}

\textbf{Probe Retrieval.}
For each shape we select a patch as a \emph{probe} (green) and highlight all patches whose cosine distance in $\mathcal{Z}$ falls below a fixed threshold (blue); all others remain gray (Fig.~\ref{fig:probe}).
On the filigree, two probe choices retrieve the four equivalent corners and the eight corners, respectively, confirming that $\mathcal{Z}$ captures both 4-fold and 8-fold rotational structure.
On the chair, probing a leg and a leg tip each retrieves all spatial counterparts across the shape.
The faucet also reveals a failure mode: probing a half-round surface patch retrieves nearly every patch, because the faucet body is topologically a pipe and most patches share the same local geometry.
This shows that cosine similarity alone is insufficient, and motivates the ICP-based verification (Sec.~\ref{sec:symmetry_detection}).

\begin{figure}[h]
    \centering
    {%
    \setlength{\tabcolsep}{0pt}%
    \newlength{\tsw}\setlength{\tsw}{0.34\columnwidth}%
    \def\chairw{0.10\columnwidth}%
    \def\faucetw{0.12\columnwidth}%
    \begin{tabular}{@{} >{\centering\arraybackslash}m{\tsw} @{\hspace{4pt}} >{\centering\arraybackslash}m{\chairw} @{\hspace{4pt}}!{\color{darkgray}\vrule}@{\hspace{4pt}} >{\centering\arraybackslash}m{\tsw} @{\hspace{4pt}} >{\centering\arraybackslash}m{\faucetw} @{}}
        \multirow{3}{\tsw}{\begin{tikzpicture}[baseline=(img.base)]
          \node[inner sep=0pt] (img) {\includegraphics[width=\tsw]{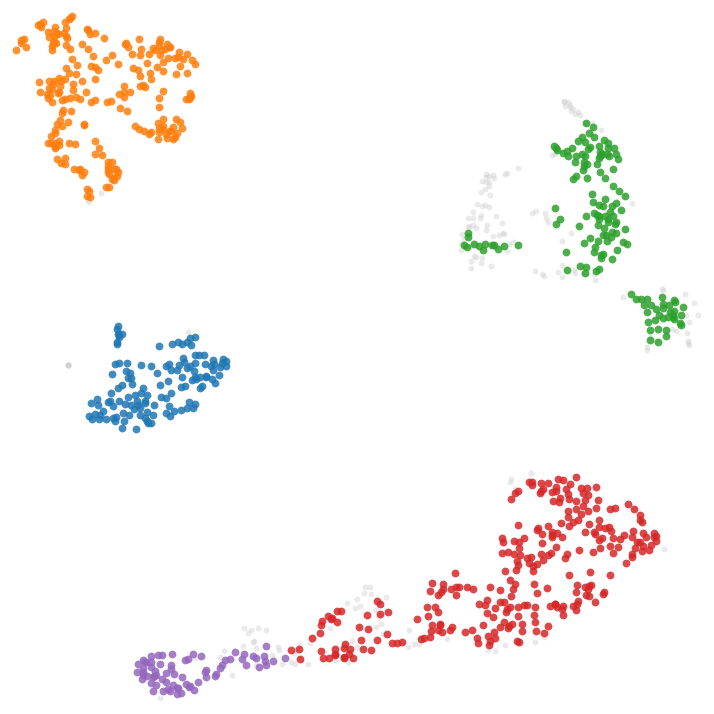}};
          \draw[black,line width=0.8pt] ($(img.center)+({-0.36\tsw},{0.35\tsw})$) circle ({0.158\tsw});
          \draw[black,line width=0.8pt] ($(img.center)+({-0.28\tsw},{-0.03\tsw})$) circle ({0.121\tsw});
          \draw[black,line width=0.8pt, rotate around={10:($(img.center)+({0.11\tsw},{-0.32\tsw})$)}] ($(img.center)+({0.11\tsw},{-0.32\tsw})$) ellipse ({0.286\tsw} and {0.187\tsw});
        \end{tikzpicture}} &
        \includegraphics[width=\chairw]{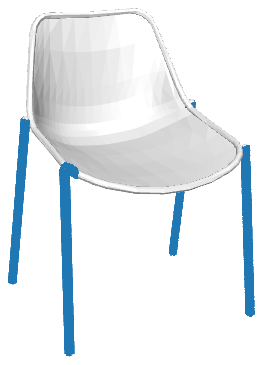} &
        \multirow{3}{\tsw}{\begin{tikzpicture}[baseline=(img.base)]
          \node[inner sep=0pt] (img) {\includegraphics[width=\tsw]{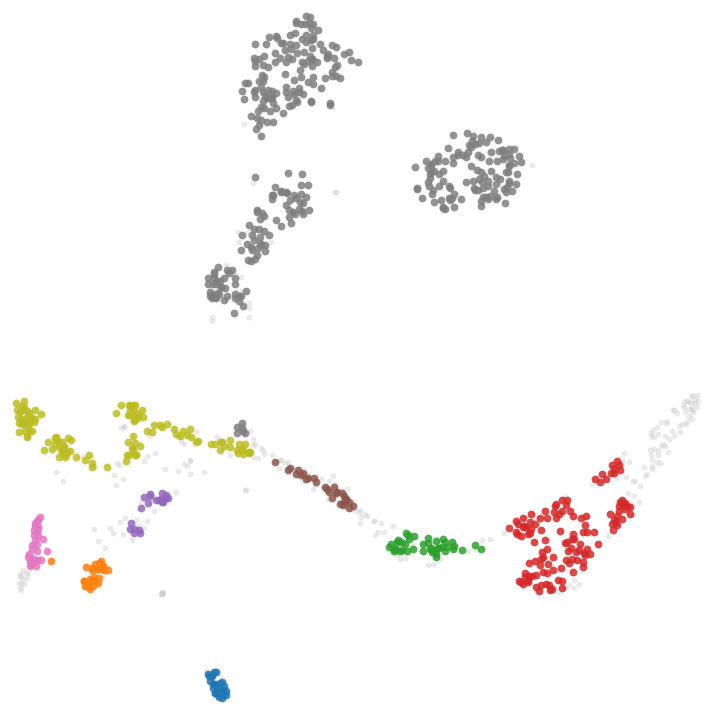}};
          \draw[black,line width=0.8pt] ($(img.center)+({-0.30\tsw},{-0.11\tsw})$) ellipse ({0.22\tsw} and {0.13\tsw});
          \draw[black,line width=0.8pt] ($(img.center)+({-0.44\tsw},{-0.26\tsw})$) circle ({0.07\tsw});
          \draw[black,line width=0.8pt] ($(img.center)+({0.105\tsw},{-0.255\tsw})$) circle ({0.084\tsw});
        \end{tikzpicture}} &
        \includegraphics[width=\faucetw]{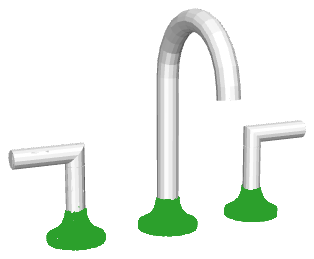} \\[2pt]
        & \includegraphics[width=\chairw]{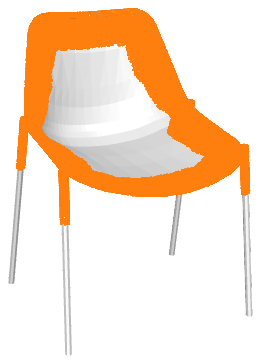} &
        & \includegraphics[width=\faucetw]{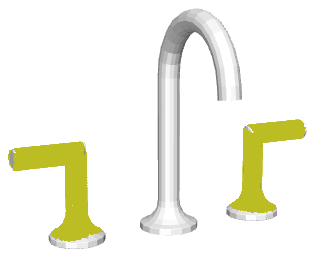} \\[2pt]
        & \includegraphics[width=\chairw]{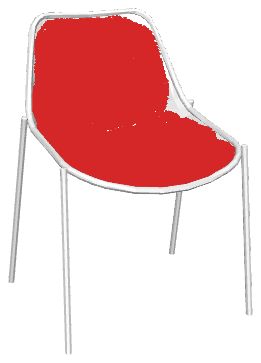} &
        & \includegraphics[width=\faucetw]{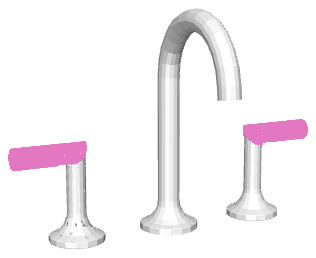}
    \end{tabular}
    }%
    \caption{
        t-SNE of patch embeddings colored by HDBSCAN cluster (small patches, gray: noise).
        Clusters close in $\mathcal{Z}$ form coherent parts on the shape (right of each plot).
        The orange chair cluster is a false positive filtered by ICP verification.
    }
    \label{fig:tsne}
\end{figure}

\textbf{t-SNE Visualization.}
Fig.~\ref{fig:tsne} projects patch embeddings into 2D via t-SNE.
Patches that are close in the embedding space $\mathcal{Z}$ form spatially coherent regions on the shape, confirming that HDBSCAN cluster membership is geometrically meaningful.
Not all clusters correspond to valid symmetry groups: the orange chair cluster (Fig.~\ref{fig:tsne}, top right) is a false positive that the ICP verification step subsequently filters out (Sec.~\ref{sec:symmetry_detection}).

\subsection{Quantitative Evaluation on SymPartNet}
\label{sec:quantitative}

\begin{figure}[t]
    \centering
    \setlength{\tabcolsep}{5pt}
    \newcommand{\sph}{0.26\columnwidth}%
    \begin{tabular}{cc @{\hspace{10pt}}!{\color{darkgray}\vrule}@{\hspace{10pt}} cc}
        \textbf{GT} & \textbf{Ours} & \textbf{GT} & \textbf{Ours} \\
        \includegraphics[height=\sph]{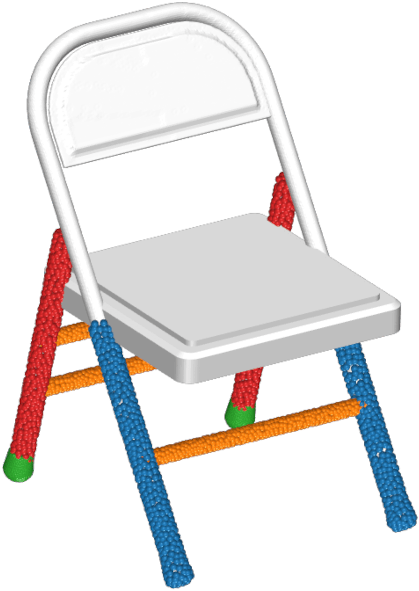} &
        \includegraphics[height=\sph]{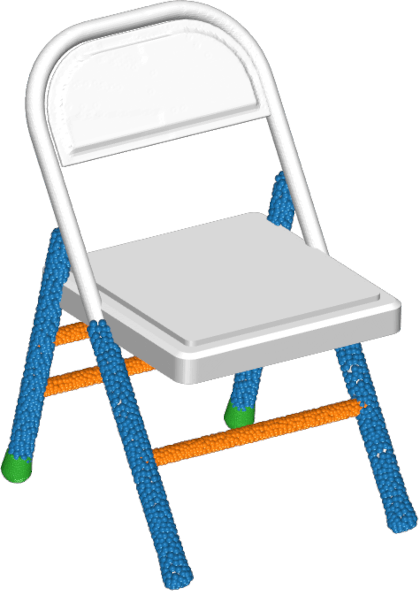} &
        \includegraphics[height=\sph]{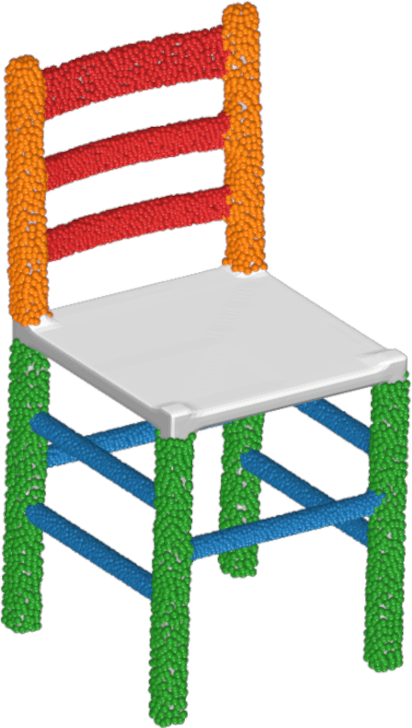} &
        \includegraphics[height=\sph]{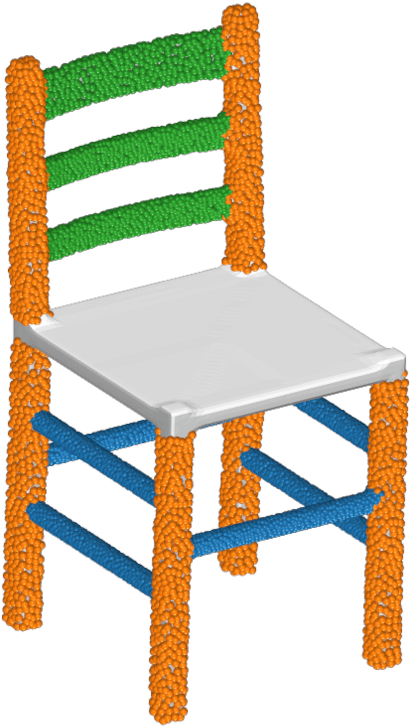} \\
    \end{tabular}
    \caption{
        Qualitative results on SymPartNet (Parts) for two chairs, each color denoting one detected symmetry group.
        SymCL groups the mutually symmetric parts (legs, cross-bars, feet) into shared clusters, closely matching the ground-truth part-symmetry annotations.
    }
    \label{fig:sympartnet_qual}
\end{figure}

\begin{figure}[t]
    \centering
    \includegraphics[width=\columnwidth]{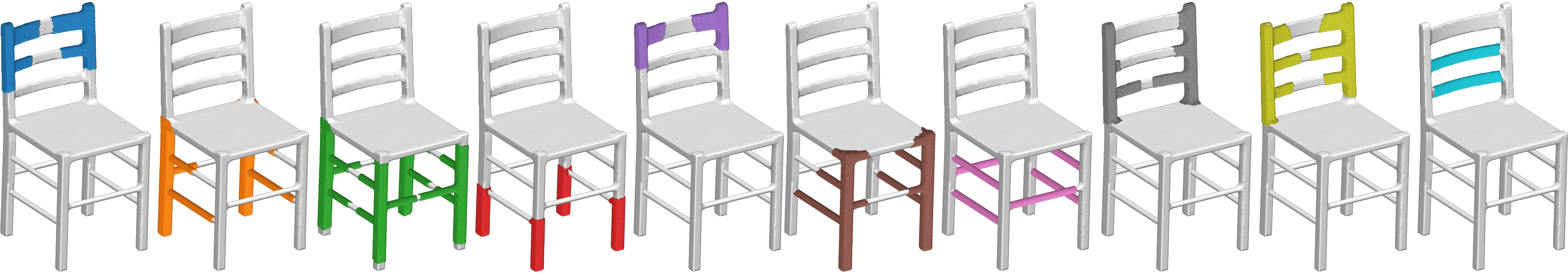}
    \caption{
        Detected symmetries on SymPartNet (Full) for chairs (SymCL).
        Each color represents a raw symmetry cluster.
    }
    \label{fig:qual_eval_epsb}
\end{figure}

We quantify detection quality on SymPartNet, our annotation of the full PartNet test set spanning 24 object categories (Sec.~\ref{sec:benchmark}).
We report two protocols: \emph{SymPartNet (Parts)}, where pre-segmented parts are given and the task is to group symmetric parts, and \emph{SymPartNet (Full)}, where only the raw model is given and symmetric regions must be discovered from scratch.
We abbreviate them as \emph{Parts} and \emph{Full}.
Table~\ref{tab:benchmark_summary} aggregates results across all categories.
SymCL retrieves symmetric structure with high coverage on both protocols (COV $0.82$ on Parts, $0.85$ on Full) at a low geometric residual (ICP $\leq 0.0026$); Fig.~\ref{fig:sympartnet_qual} shows representative Parts predictions next to the ground truth.
The drop in IoU from Parts ($0.70$) to Full ($0.32$) reflects the harder unsegmented setting: PartNet's ground-truth partition is semantically driven, whereas SymCL partitions geometrically, so predicted regions need not align with semantic part boundaries; the consistently high COV confirms that symmetric regions are nonetheless recovered (Fig.~\ref{fig:qual_eval_epsb}).
These scores nevertheless leave clear room for improvement: the benchmark is far from saturated and remains an open challenge for future methods.

\begin{table}[h]
    \centering
    \input{tab/benchmark_summary.tex}
    \caption{
        Quantitative results on SymPartNet, aggregated over all 24 PartNet categories (test split), under automatically derived symmetry annotations.
        Parts evaluates on pre-segmented parts, Full on raw models.
    }
    \label{tab:benchmark_summary}
\end{table}

\subsection{Generalization to Unseen Geometry}
\label{sec:generalization}

Since SymCL learns purely geometric, patch-level features, it generalizes to shape categories not seen during PartNet training.

\textbf{Out-of-Distribution Shapes.}
Fig.~\ref{fig:region_growing} shows SymCL applied to shapes outside PartNet.
In all cases, recurring sub-structures are correctly grouped and region growing expands the detected patches to maximal symmetric partitions.
The same shape can yield multiple valid symmetry hypotheses simultaneously (e.g.\ global left-right reflection and local rotational leg repetition of a chair coexist as distinct clusters).

\subsection{Comparative Evaluation}
\label{sec:comparison}

We compare SymCL against the two most relevant competitors on representative shapes spanning reflective, rotational, and translational symmetry: Mitra et al.~\cite{mitra2006partial} (classical pairwise voting) and Je et al.~\cite{je2024robust} (reflection-plane optimization).
A shared quantitative comparison is not directly possible: Je et al.\ output reflection planes and Mitra et al.\ output pairwise transformations, whereas SymCL outputs multi-instance region clusters.
We therefore compare qualitatively, selecting shapes whose structure each method is designed to address.
The three methods differ in where symmetry is decided.
Je et al.\ are top-down: a global plane hypothesis is fitted first, and supporting regions are derived from it.
Mitra et al.\ are bottom-up but reason in an explicit transformation space, voting with pairwise point correspondences and clustering the votes into candidate symmetries; this recovers individual transformations yet aggregates pairwise evidence and struggles to group all instances of a recurring structure.
SymCL is also bottom-up but clusters patches in a learned invariant feature space, so multi-instance symmetry groups emerge directly, without enumerating explicit transformations or committing to a symmetry type or arity.

\begin{figure*}[h]
    \centering
    \setlength{\tabcolsep}{1pt}
    \newcommand{\sw}{0.094\textwidth}
    \resizebox{\textwidth}{!}{%
    \begin{tabular}{>{\centering\arraybackslash}m{1.5em} @{\hspace{4pt}} *{3}{>{\centering\arraybackslash}m{\sw}} @{\hspace{4pt}}!{\color{darkgray}\vrule}@{\hspace{4pt}} *{3}{>{\centering\arraybackslash}m{\sw}} @{\hspace{4pt}}!{\color{darkgray}\vrule}@{\hspace{4pt}} *{3}{>{\centering\arraybackslash}m{\sw}}}
        & \multicolumn{3}{c}{\textbf{Mitra et al.}}
        & \multicolumn{3}{c}{\textbf{Je et al.}}
        & \multicolumn{3}{c}{\textbf{Ours}} \\[3pt]
        \rotatebox{90}{\textbf{Filigree}} &
            \includegraphics[width=\sw]{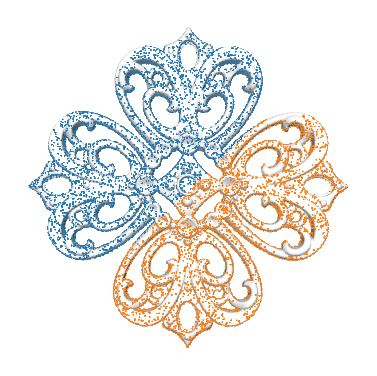} &
            \includegraphics[width=\sw]{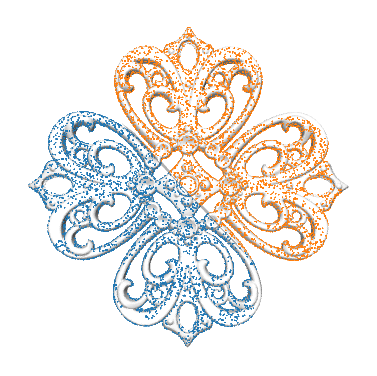} &
            \includegraphics[width=\sw]{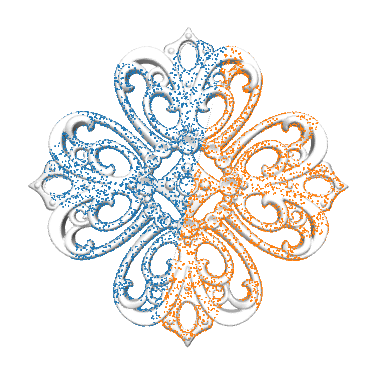} &
            \includegraphics[width=\sw]{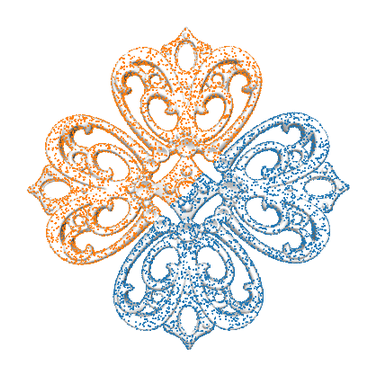} &
            \includegraphics[width=\sw]{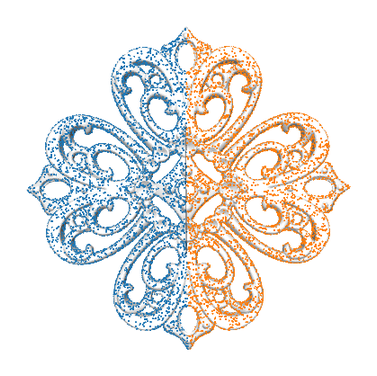} &
            \includegraphics[width=\sw]{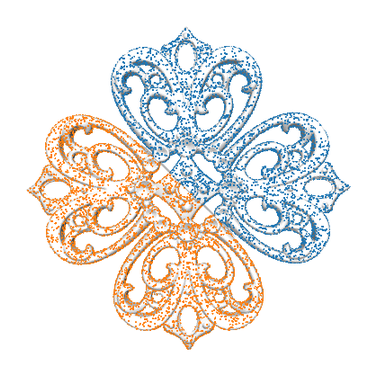} &
            \includegraphics[width=\sw]{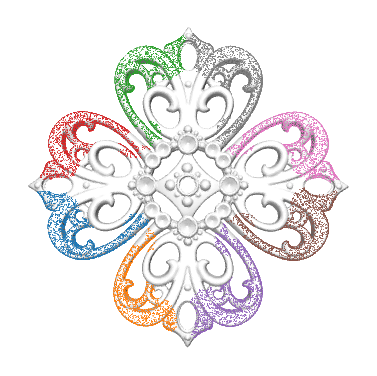} &
            \includegraphics[width=\sw]{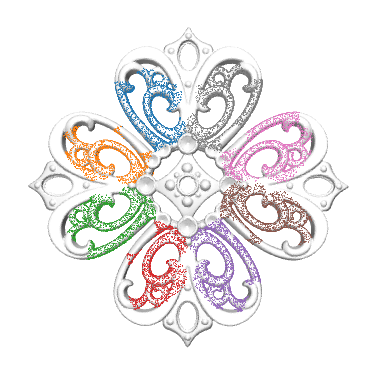} &
            \includegraphics[width=\sw]{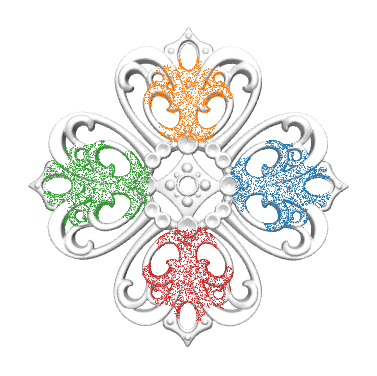} \\[3pt]
        \rotatebox{90}{\textbf{Chair}} &
            \includegraphics[width=\sw]{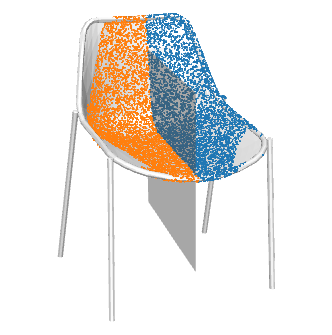} &
            \includegraphics[width=\sw]{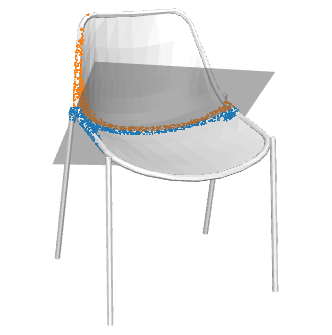} &
            \includegraphics[width=\sw]{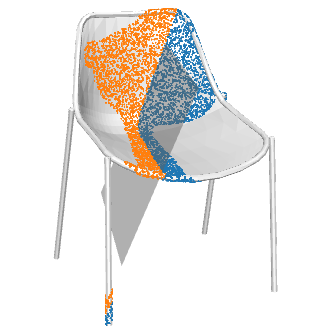} &
            \includegraphics[width=\sw]{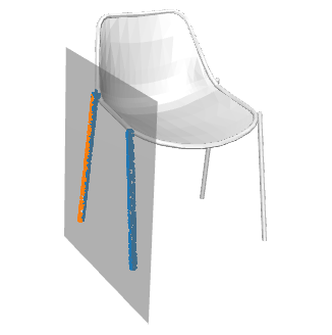} &
            \includegraphics[width=\sw]{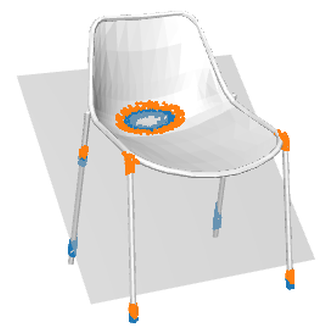} &
            \includegraphics[width=\sw]{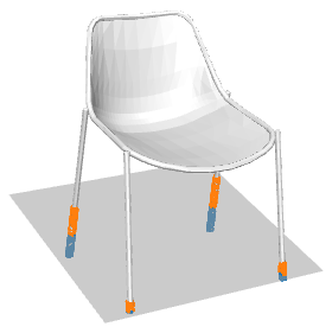} &
            \includegraphics[width=\sw]{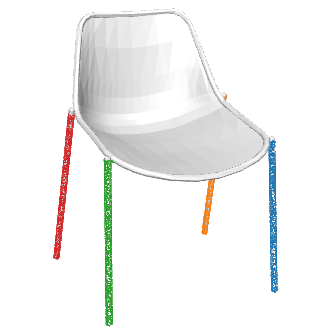} &
            \includegraphics[width=\sw]{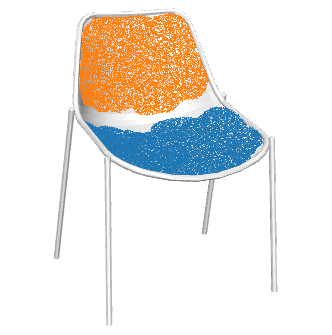} &
            \includegraphics[width=\sw]{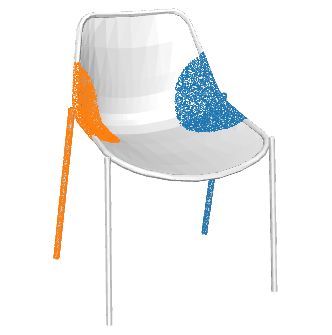} \\[3pt]
        \rotatebox{90}{\textbf{Faucet}} &
            \includegraphics[width=\sw]{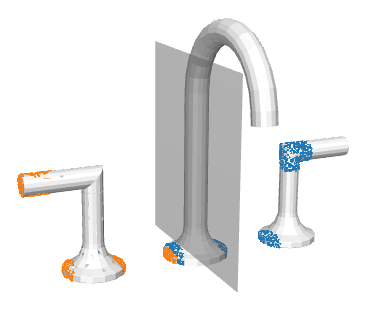} &
            \includegraphics[width=\sw]{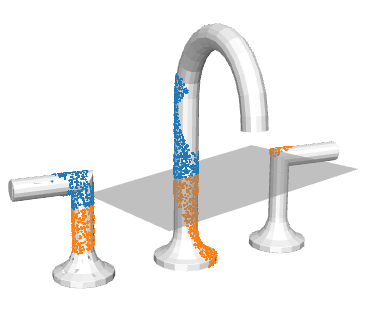} &
            \includegraphics[width=\sw]{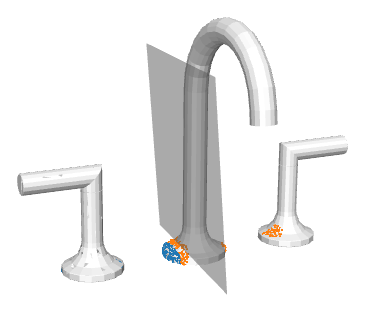} &
            \includegraphics[width=\sw]{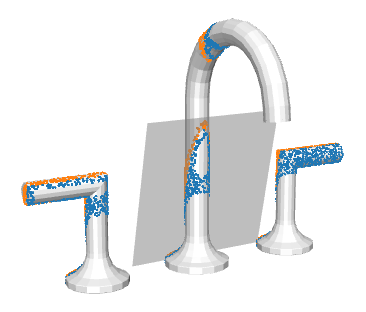} &
            \includegraphics[width=\sw]{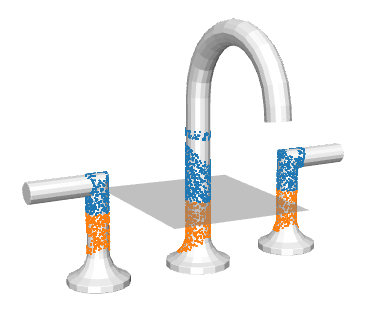} &
            \includegraphics[width=\sw]{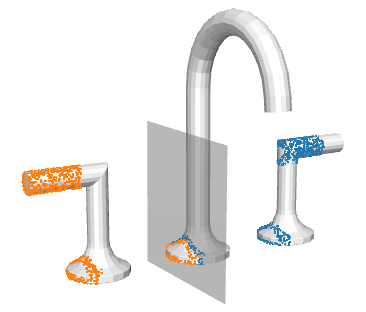} &
            \includegraphics[width=\sw]{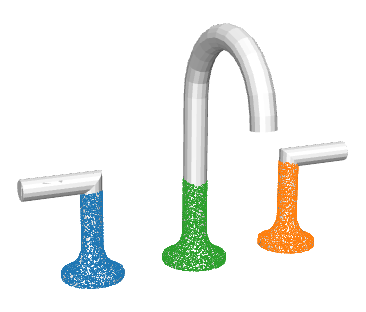} &
            \includegraphics[width=\sw]{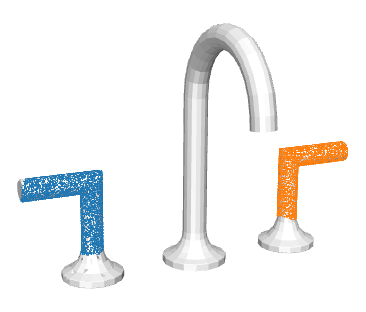} &
            \includegraphics[width=\sw]{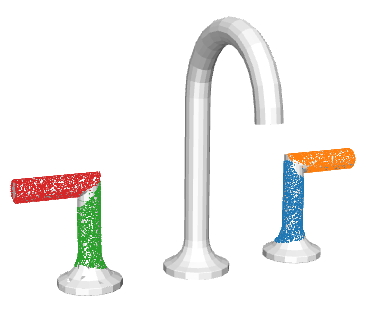} \\
    \end{tabular}}
    \caption{
        Qualitative comparison on three shapes.
        \textbf{Filigree}: a delicate but topologically simple shape; both competitors recover only 2-fold symmetries, while SymCL detects higher-order 4-fold and 8-fold rotational symmetries.
        \textbf{Chair}: SymCL recovers both 2-fold and 4-fold symmetries; the symmetry planes proposed by Je et al.\ appear largely arbitrary, and Mitra et al.\ finds only one meaningful correspondence.
        \textbf{Faucet}: SymCL detects 2-fold reflectional, 3-fold translational, and 4-fold combined symmetries; Mitra et al.\ recovers only the global left-right reflection plane with very low support; Je et al.\ proposes three planes that are too noisy to recover their supporting points reliably.
        Blue and orange points indicate the two sides of each detected symmetry; each color in the Ours column denotes a distinct symmetric component.
    }
    \label{fig:comparison}
\end{figure*}

For Je et al.\ we used the official implementation with the default parameters provided by the authors.
No publicly available implementation of Mitra et al.\ exists; we therefore reimplemented the method to the best of our ability following the original paper, noting that several technical details were omitted in the publication.
Since Mitra et al.'s algorithm requires per-shape parameter tuning, we performed a grid search over the key parameters and report the best-performing configuration for each shape individually.
Notably, across all shapes tested, Mitra et al.'s method produced rotational symmetry detections only for two simple cases (a bow and a star); all other shapes yielded reflection symmetries exclusively.
Figure~\ref{fig:comparison} shows three representative objects covering the complementary strengths and limitations of each approach.

\textbf{Filigree.}
Despite its intricate appearance, the filigree is a topologically simple shape with clearly repeated motifs arranged in higher-order rotational configurations.
Both competitors recover only 2-fold reflection symmetries, missing the dominant structure of the shape entirely.
SymCL detects 4-fold and 8-fold rotational symmetry groups, recovering all recurring motifs as coherent clusters.

\textbf{Chair.}
A chair with four identical legs provides both 2-fold and 4-fold symmetries.
The symmetry planes proposed by Je et al.\ appear largely arbitrary, capturing neither the global reflection nor the leg repetition reliably.
Mitra et al.\ recovers the global left-right symmetry plane.
SymCL detects the reflective symmetry between seat and backrest, the 4-fold rotational symmetry of the legs, and partially the global left-right symmetry of the chair.

\textbf{Faucet.}
The faucet exhibits a richer symmetry structure: 2-fold reflectional symmetry, 3-fold translational repetition in the base, and 4-fold combined symmetries.
Mitra et al.\ recovers only the global left-right reflection plane, with very low supporting point coverage.
Je et al.\ proposes three symmetry planes, but averaging and clustering in the fitting stage produces planes too noisy to identify coherent supporting regions.
SymCL is able to detect all three symmetry types in a single forward pass.

Overall, the detected symmetry planes of classical and optimization-based methods are highly sensitive to parameter choices, prone to noise, and fundamentally limited to 2-fold partial symmetries.
SymCL overcomes some of these limitations on the presented samples, but reliable partial symmetry detection remains an open challenge.

\subsection{Limitations and Future Work}

A single threshold $\delta_{sim}$ for filtering symmetry hypotheses remains the only hand-tuned parameter, and the region growing expands patches equidistantly; removing the threshold and adaptive per-patch growing rates are directions for future work.
Finally, SymPartNet's ground truth inherits PartNet's semantic segmentation and thus favors semantically driven methods over purely geometric ones such as ours; a purely geometry-driven benchmark, and a broader evaluation of competing methods on it, are left to future work.

\section{Conclusion}
\label{sec:conclusion}

We presented SymCL, a self-supervised framework for partial extrinsic symmetry detection that requires no symmetry annotations and makes no prior commitment to symmetry type.
By clustering $E(3)$-invariant geodesic-patch embeddings, it recovers multi-instance symmetry groups of rotational, translational, and reflective structure in a single forward pass, bypassing the pairwise voting of classical methods.
Qualitative comparisons show that SymCL recovers higher-order symmetries that plane- and voting-based competitors systematically miss, while generalizing to shape categories unseen during training.
To evaluate partial extrinsic symmetry at scale, we further introduced SymPartNet, a benchmark annotating all PartNet categories with part-level symmetry relations.
Our code, the reimplementation of Mitra et al.~\cite{mitra2006partial}, and the benchmark will be made publicly available upon publication.


\vspace{-8pt}
\bibliographystyle{eg-alpha}
\bibliography{main}

\end{document}

%% file: tab/benchmark_summary.tex
\begin{tabular}{lcccc}
  \toprule
  \multicolumn{5}{c}{\textbf{SymPartNet}}                      \\
                  & ICP (↓) & IoU (↑) & COV (↑) & \#Models \\
  \midrule
  \textbf{Parts}  & 0.0026  & 0.7000  & 0.8176  & 3071 \\
  \textbf{Full}   & 0.0025  & 0.3186  & 0.8522  & 3155 \\
  \bottomrule
\end{tabular}

%% file: main.bbl
\newcommand{\etalchar}[1]{$^{#1}$}
\begin{thebibliography}{\uppercase{MPWC13}}

\bibitem[ADD{\etalchar{*}}22]{afham2022crosspoint}
\textsc{Afham M., Dissanayake I., Dissanayake D., Dharmasiri A., Thilakarathna
  K., Rodrigo R.}:
\newblock Crosspoint: Self-supervised cross-modal contrastive learning for 3d
  point cloud understanding.
\newblock In \emph{IEEE/CVF Conference on Computer Vision and Pattern
  Recognition (CVPR)} (2022), pp.~9902--9912.

\bibitem[Ata85]{atallah1985symmetry}
\textsc{Atallah M.~J.}:
\newblock On symmetry detection.
\newblock \emph{IEEE Transactions on Computers 34}, 7 (1985), 663--666.

\bibitem[BBW{\etalchar{*}}09]{bokeloh2009symmetry}
\textsc{Bokeloh M., Berner A., Wand M., Seidel H.-P., Schilling A.}:
\newblock Symmetry detection using feature lines.
\newblock In \emph{Computer Graphics Forum} (2009), vol.~28, Wiley Online
  Library, pp.~697--706.

\bibitem[BM92]{besl1992method}
\textsc{Besl P.~J., McKay N.~D.}:
\newblock A method for registration of 3-d shapes.
\newblock \emph{IEEE Transactions on Pattern Analysis and Machine Intelligence
  14}, 2 (1992), 239--256.

\bibitem[CKNH20]{chen2020simple}
\textsc{Chen T., Kornblith S., Norouzi M., Hinton G.}:
\newblock A simple framework for contrastive learning of visual
  representations.
\newblock In \emph{International Conference on Machine Learning (ICML)} (2020),
  pp.~1597--1607.

\bibitem[CWW17]{crane2017heat}
\textsc{Crane K., Weischedel C., Wardetzky M.}:
\newblock The heat method for distance computation.
\newblock \emph{Communications of the ACM 60}, 11 (2017), 90--99.

\bibitem[DBI18]{deng2018ppffoldnet}
\textsc{Deng H., Birdal T., Ilic S.}:
\newblock Ppf-foldnet: Unsupervised learning of rotation invariant 3d local
  descriptors.
\newblock In \emph{European Conference on Computer Vision (ECCV)} (2018),
  pp.~602--618.

\bibitem[DLD{\etalchar{*}}21]{deng2021vector}
\textsc{Deng C., Litany O., Duan Y., Poulenard A., Tagliasacchi A., Guibas
  L.~J.}:
\newblock Vector neurons: A general framework for so(3)-equivariant networks.
\newblock In \emph{International Conference on Computer Vision (ICCV)} (2021),
  pp.~12200--12209.

\bibitem[FSG17]{fan2017point}
\textsc{Fan H., Su H., Guibas L.~J.}:
\newblock A point set generation network for 3d object reconstruction from a
  single image.
\newblock In \emph{IEEE Conference on Computer Vision and Pattern Recognition
  (CVPR)} (2017), pp.~605--613.

\bibitem[GCO06]{gal2006salient}
\textsc{Gal R., Cohen-Or D.}:
\newblock Salient geometric features for partial shape matching and similarity.
\newblock \emph{ACM Transactions on Graphics (TOG) 25}, 1 (2006), 130--150.

\bibitem[GYW{\etalchar{*}}19]{gao2019sdm}
\textsc{Gao L., Yang J., Wu T., Yuan Y.-J., Fu H., Lai Y.-K., Zhang H.}:
\newblock Sdm-net: Deep generative network for structured deformable mesh.
\newblock \emph{ACM Transactions on Graphics 38}, 6 (2019), 1--15.

\bibitem[GZM{\etalchar{*}}20]{gao2020prs}
\textsc{Gao L., Zhang L.-X., Meng H.-Y., Ren Y.-H., Lai Y.-K., Kobbelt L.}:
\newblock Prs-net: Planar reflective symmetry detection net for 3d models.
\newblock \emph{IEEE Transactions on Visualization and Computer Graphics 27}, 6
  (2020), 3007--3018.

\bibitem[JL19]{ji2019fast}
\textsc{Ji P., Liu X.}:
\newblock A fast and efficient 3d reflection symmetry detector based on neural
  networks.
\newblock \emph{Multimedia Tools and Applications 78} (2019), 35471--35492.

\bibitem[JLY{\etalchar{*}}24]{je2024robust}
\textsc{Je J., Liu J., Yang G., Deng B., Cai S., Wetzstein G., Litany O.,
  Guibas L.~J.}:
\newblock Robust symmetry detection via riemannian langevin dynamics.
\newblock In \emph{ACM SIGGRAPH Asia 2024 Conference Papers} (2024), pp.~1--11.

\bibitem[KCD{\etalchar{*}}04]{kazhdan2004reflective}
\textsc{Kazhdan M., Chazelle B., Dobkin D., Funkhouser T., Rusinkiewicz S.}:
\newblock A reflective symmetry descriptor for 3d models.
\newblock \emph{Algorithmica 38} (2004), 201--225.

\bibitem[LCDF10]{lipman2010symmetry}
\textsc{Lipman Y., Chen X., Daubechies I., Funkhouser T.}:
\newblock Symmetry factored embedding and distance.
\newblock \emph{ACM Transactions on Graphics 29}, 4 (July 2010).

\bibitem[LHTR25]{li2025symmetry}
\textsc{Li X., Huang Z., Thai A., Rehg J.~M.}:
\newblock Symmetry strikes back: From single-image symmetry detection to 3d
  generation.
\newblock In \emph{IEEE/CVF Conference on Computer Vision and Pattern
  Recognition (CVPR)} (2025).

\bibitem[LJYL16]{li2016efficient}
\textsc{Li B., Johan H., Ye Y., Lu Y.}:
\newblock Efficient 3d reflection symmetry detection: A view-based approach.
\newblock \emph{Graphical Models 83} (2016), 2--14.

\bibitem[LXC{\etalchar{*}}17]{li2017grass}
\textsc{Li J., Xu K., Chaudhuri S., Yumer E., Zhang H., Guibas L.}:
\newblock Grass: Generative recursive autoencoders for shape structures.
\newblock \emph{ACM Transactions on Graphics 36}, 4 (2017), 1--14.

\bibitem[LZL{\etalchar{*}}23]{li2023e3sym}
\textsc{Li R.-W., Zhang L.-X., Li C., Lai Y.-K., Gao L.}:
\newblock E3sym: Leveraging e(3) invariance for unsupervised 3d planar
  reflective symmetry detection.
\newblock In \emph{International Conference on Computer Vision (ICCV)} (2023),
  pp.~14543--14553.

\bibitem[MBB10]{mitra2010intrinsic}
\textsc{Mitra N.~J., Bronstein A., Bronstein M.}:
\newblock Intrinsic regularity detection in 3d geometry.
\newblock In \emph{European Conference on Computer Vision (ECCV)} (2010),
  Springer, pp.~398--410.

\bibitem[MGP06]{mitra2006partial}
\textsc{Mitra N.~J., Guibas L.~J., Pauly M.}:
\newblock Partial and approximate symmetry detection for 3d geometry.
\newblock \emph{ACM Transactions on Graphics 25}, 3 (2006), 560--568.

\bibitem[MGY{\etalchar{*}}19]{mo2019structurenet}
\textsc{Mo K., Guerrero P., Yi L., Su H., Wonka P., Mitra N., Guibas L.}:
\newblock Structurenet: Hierarchical graph networks for 3d shape generation.
\newblock \emph{ACM Transactions on Graphics 38}, 6 (2019), Article 242.

\bibitem[MHA17]{mcinnes2017hdbscan}
\textsc{McInnes L., Healy J., Astels S.}:
\newblock hdbscan: Hierarchical density based clustering.
\newblock \emph{The Journal of Open Source Software 2}, 11 (2017), 205.

\bibitem[MPWC13]{mitra2013symmetry}
\textsc{Mitra N.~J., Pauly M., Wand M., Ceylan D.}:
\newblock Symmetry in 3d geometry: Extraction and applications.
\newblock In \emph{Computer Graphics Forum} (2013), vol.~32, Wiley Online
  Library, pp.~1--23.

\bibitem[MSHS06]{martinet2006accurate}
\textsc{Martinet A., Soler C., Holzschuch N., Sillion F.~X.}:
\newblock Accurate detection of symmetries in 3d shapes.
\newblock \emph{ACM Transactions on Graphics 25}, 2 (2006), 439--464.

\bibitem[MZC{\etalchar{*}}19]{mo2019partnet}
\textsc{Mo K., Zhu S., Chang A.~X., Yi L., Tripathi S., Guibas L.~J., Su H.}:
\newblock Partnet: A large-scale benchmark for fine-grained and hierarchical
  part-level 3d object understanding.
\newblock In \emph{IEEE/CVF Conference on Computer Vision and Pattern
  Recognition (CVPR)} (2019), pp.~909--918.

\bibitem[NR18]{nagar2018fast}
\textsc{Nagar R., Raman S.}:
\newblock Fast and accurate intrinsic symmetry detection.
\newblock In \emph{European Conference on Computer Vision (ECCV)} (2018),
  pp.~417--434.

\bibitem[OSG08]{ovsjanikov2008global}
\textsc{Ovsjanikov M., Sun J., Guibas L.}:
\newblock Global intrinsic symmetries of shapes.
\newblock In \emph{Computer Graphics Forum} (2008), vol.~27, Wiley Online
  Library, pp.~1341--1348.

\bibitem[PMW{\etalchar{*}}08]{pauly2008discovering}
\textsc{Pauly M., Mitra N.~J., Wallner J., Pottmann H., Guibas L.}:
\newblock Discovering structural regularity in {3D} geometry.
\newblock \emph{ACM Transactions on Graphics 27}, 3 (2008), \#43, 1--11.

\bibitem[PSG{\etalchar{*}}06]{podolak2006planar}
\textsc{Podolak J., Shilane P., Golovinskiy A., Rusinkiewicz S., Funkhouser
  T.}:
\newblock A planar-reflective symmetry transform for {3D} shapes.
\newblock \emph{ACM Transactions on Graphics 25}, 3 (July 2006).

\bibitem[QGL{\etalchar{*}}22]{qiao2022learning}
\textsc{Qiao Y.-L., Gao L., Liu S.-Z., Liu L., Lai Y.-K., Chen X.}:
\newblock Learning-based intrinsic reflectional symmetry detection.
\newblock \emph{IEEE Transactions on Visualization and Computer Graphics 29}, 9
  (2022).

\bibitem[QYSG17]{qi2017pointnet++}
\textsc{Qi C.~R., Yi L., Su H., Guibas L.~J.}:
\newblock Pointnet++: Deep hierarchical feature learning on point sets in a
  metric space.
\newblock \emph{Advances in Neural Information Processing Systems 30} (2017).

\bibitem[SC{\etalchar{*}}19]{geometrycentral}
\textsc{Sharp N., Crane K., et~al.}:
\newblock Geometrycentral: A modern c++ library of data structures and
  algorithms for geometry processing.
\newblock \url{https://geometry-central.net/}, 2019.

\bibitem[SGS14]{sipiran2014approximate}
\textsc{Sipiran I., Gregor R., Schreck T.}:
\newblock Approximate symmetry detection in partial 3d meshes.
\newblock In \emph{Computer Graphics Forum} (2014), vol.~33, Wiley Online
  Library, pp.~131--140.

\bibitem[Soh16]{sohn2016improved}
\textsc{Sohn K.}:
\newblock Improved deep metric learning with multi-class n-pair loss objective.
\newblock \emph{Advances in Neural Information Processing Systems (NeurIPS) 29}
  (2016).

\bibitem[SRF{\etalchar{*}}23]{sipiran2023shrec}
\textsc{Sipiran I., Romanengo C., Falcidieno B., Biasotti S., Arvanitis G.,
  Chen C., Fotis V., He J., Lv X., Moustakas K., Peng S., Romanelis I., Sun W.,
  Vlachos C., Wu Z., Xie Q.}:
\newblock {SHREC} 2023: Detection of symmetries on 3d point clouds representing
  simple shapes.
\newblock In \emph{Eurographics Workshop on 3D Object Retrieval (3DOR) - Short
  Papers} (2023), The Eurographics Association.

\bibitem[WH17]{wang2017group}
\textsc{Wang H., Huang H.}:
\newblock Group representation of global intrinsic symmetries.
\newblock \emph{Computer Graphics Forum 36}, 7 (2017), 51--61.

\bibitem[XGG{\etalchar{*}}20]{xie2020pointcontrast}
\textsc{Xie S., Gu J., Guo D., Qi C.~R., Guibas L., Litany O.}:
\newblock Pointcontrast: Unsupervised pre-training for 3d point cloud
  understanding.
\newblock In \emph{European Conference on Computer Vision (ECCV)} (2020),
  pp.~574--591.

\bibitem[XZJ{\etalchar{*}}12]{xu2012multi}
\textsc{Xu K., Zhang H., Jiang W., Dyer R., Cheng Z., Liu L., Chen B.}:
\newblock Multi-scale partial intrinsic symmetry detection.
\newblock \emph{ACM Transactions on Graphics 31}, 6 (2012), 1--11.

\bibitem[XZT{\etalchar{*}}09]{xu2009partial}
\textsc{Xu K., Zhang H., Tagliasacchi A., Liu L., Li G., Meng M., Xiong Y.}:
\newblock Partial intrinsic reflectional symmetry of 3d shapes.
\newblock \emph{ACM Transactions on Graphics 28}, 5 (2009), 138:1--138:10.

\bibitem[YLZ{\etalchar{*}}19]{yu2019partnet}
\textsc{Yu F., Liu K., Zhang Y., Zhu C., Xu K.}:
\newblock Partnet: A recursive part decomposition network for fine-grained and
  hierarchical shape segmentation.
\newblock In \emph{IEEE/CVF Conference on Computer Vision and Pattern
  Recognition (CVPR)} (2019), pp.~9491--9500.

\end{thebibliography}
